\documentclass[preprint, review, 3p, authoryear, times]{elsarticle}

\usepackage{enumitem}
\usepackage{placeins}
\usepackage{caption}
\usepackage{float} 
\usepackage{graphicx}
\usepackage{url}
\usepackage{color, soul}
\usepackage{bm}
\usepackage{amsmath} 
\usepackage{lineno}

\begin{document}


\title{
Explainable AI for Multivariate Time Series Pattern Exploration: Latent Space Visual Analytics with Temporal Fusion Transformer and Variational Autoencoders in Power Grid Event Diagnosis} 



\tnotetext[t1]{This manuscript has been co-authored by UT-Battelle, LLC, under contract DE-AC05-00OR22725 with the US Department of Energy (DOE). The US government retains and the publisher, by accepting the article for publication, acknowledges that the US government retains a nonexclusive, paid-up, irrevocable, worldwide license to publish or reproduce the published form of this manuscript, or allow others to do so, for US government purposes. DOE will provide public access to these results of federally sponsored research in accordance with the DOE Public Access Plan (http://energy.gov/downloads/doe-public-access-plan).}
\author[CSED]{Haowen Xu \corref{cor1}}\ead{xuh4@ornl.gov}
\author[GCSG]{Ali Boyaci} \ead{boyacia@ornl.gov}
\author[GCG]{Jianming (Jamie) Lian} \ead{}
\author[GCSG]{Aaron Wilson} \ead{}

\cortext[cor1]{Corresponding author.}

\address[CSED]{Computational Urban Sciences Group, Oak Ridge National Laboratory, Oak Ridge, TN 37830, USA}
\address[GCSG]{Grid Communications and Security Group, Oak Ridge National Laboratory, Oak Ridge, TN 37830, USA}
\address[GCG]{Grid-interactive Controls Group, Oak Ridge, TN 37830, USA} 
 

\makeatletter
\newcommand{\printfnsymbol}[1]{%
  \textsuperscript{\@fnsymbol{#1}}%
}

\newcommand*{\MyIndent}{\hspace*{0.5cm}}%


\begin{abstract} \label{sec:abstract}
Detecting and analyzing complex patterns in multivariate time-series data is crucial for decision-making in urban and environmental system operations. However, challenges arise from the high dimensionality, intricate complexity, and interconnected nature of complex patterns, which hinder the understanding of their underlying physical processes. Existing AI methods often face limitations in interpretability, computational efficiency, and scalability, reducing their applicability in real-world scenarios. This paper proposes a novel visual analytics framework that integrates two generative AI models, Temporal Fusion Transformer (TFT) and Variational Autoencoders (VAEs), to reduce complex patterns into lower-dimensional latent spaces and visualize them in 2D using dimensionality reduction techniques such as PCA, t-SNE, and UMAP with DBSCAN. These visualizations, presented through coordinated and interactive views and tailored glyphs, enable intuitive exploration of complex multivariate temporal patterns, identifying patterns’ similarities and uncover their potential correlations for a better interpretability of the AI outputs. The framework is demonstrated through a case study on power grid signal data, where it identifies multi-label grid event signatures, including faults and anomalies with diverse root causes. Additionally, novel metrics and visualizations are introduced to validate the models and evaluate the performance, efficiency, and consistency of latent maps generated by TFT and VAE under different configurations. These analyses provide actionable insights for model parameter tuning and reliability improvements. Comparative results highlight that TFT achieves shorter run times and superior scalability to diverse time-series data shapes compared to VAE. This work advances fault diagnosis in multivariate time series, fostering explainable AI to support critical system operations.

\end{abstract}

\maketitle


\section{Introduction}
\label{Introduction}
Exploring patterns in multivariate time-series data is challenging due to the complexity, heterogeneity, and high dimensionality inherent in many environmental and urban datasets \citep{catley2008multi}. These datasets often involve multiple interdependent variables measured over time, interacting in non-linear ways and across different time scales \citep{goswami2019brief}. Temporal dependencies, varying trends, noise, missing data points, and irregular sampling intervals further complicate the identification of meaningful patterns, which may correspond to specific physical phenomena or events in real-world systems \citep{weerakody2021review, sun2020review}.
To address these challenges, a range of AI-driven methods has been developed to enhance the classification and clustering of multivariate time-series data \citep{rojat2021explainable}. Traditional machine learning approaches, such as Principal Component Analysis (PCA) \citep{li2019multivariate} and Self-Organizing Maps (SOM) \citep{d2014wavelet}, have been widely applied. More recently, advanced deep learning models have demonstrated significant capabilities. Recurrent Neural Networks (RNNs) \citep{ienco2020deep, wang2016effective} and Long Short-Term Memory (LSTM) networks \citep{karim2019multivariate, hong2017multivariate} excel in capturing temporal dependencies, while Convolutional Neural Networks (CNNs) effectively identify spatial and temporal patterns \citep{liu2018time, tripathi2020multivariate}. Generative algorithms, particularly Transformer models \citep{liu2021gated, zuo2023svp, le2024shapeformer, zerveas2021transformer}, have emerged as powerful tools for modeling long-range dependencies in sequential data, making them highly suitable for multivariate time-series analysis \citep{zhou2021informer}. Variational Autoencoders (VAEs) are also gaining traction for their ability to handle missing data imputation and anomaly detection \citep{graving2020vae, lafabregue2022end, pham2022mst, yokkampon2022robust}. These deep generative models effectively learn data distributions in high-dimensional spaces by mapping them to lower-dimensional latent spaces, enabling more accurate and efficient analysis.

Despite advancements in AI-based methods, human-in-the-loop (HitL) approaches remain essential for interpreting and validating AI-generated insights \citep{kumar2024applications, gomez2024human}. Human expertise ensures these patterns are relevant to specific domains, such as power system operations, where grid event data—often comprised of multivariate time-series recordings such as voltage, frequency, and current—requires careful interpretation to maintain reliable and efficient operations \citep{tocchetti2022role, wilson2024grid}. These signatures often indicate system inefficiencies or disruptions, such as faults, generation/load imbalance, and inverter-induced oscillations, which can significantly impact grid reliability and stability if not addressed \citep{rivas2020faults, das2024modeling}. In addition, developing efficient AI-powered methods to uncover potential connections and interactions between different types of grid disturbances can enhance researchers' understanding of system complexity, enabling the development of more effective operational practices. While AI can detect these disturbances, human involvement is critical for contextualizing and applying these insights in real-world urban power system operations. This interplay highlights the importance of explainable AI (XAI) methods, which enhance transparency and empower operators to act confidently on AI-generated insights \citep{nallakaruppan2024advancing}. Visual analytics further strengthens human-AI collaboration by enabling intuitive understanding of deep learning models through interactive visualizations, emphasizing the need to integrate visual analytics into XAI methodologies \citep{niloofar2023general, xu2024leveraging}.

Addressing the challenges of analyzing complex patterns in multivariate time series data requires innovative approaches that bridge the gap between advanced deep learning models and interpretability. This paper introduces a visual analytics framework that reduces and represents these patterns in lower-dimensional latent spaces using the Temporal Fusion Transformer (TFT) and Variational Autoencoder (VAE) approaches. Dimensionality reduction techniques such as PCA, t-SNE, and UMAP are employed to visualize these representations in 2D as latent vector maps, enabling intuitive exploration of data patterns. Distance metrics are used to uncover correlations and topological relationships between patterns, enhancing the interpretability of generative models for exploratory fault pattern analysis. To ensure robust and efficient performance, the framework integrates evaluation metrics and advanced visualizations, enabling comparative analysis and validation of TFT- and VAE-based methods through domain knowledge. These tools facilitate parameter tuning, improving model reliability and scalability. 

A case study on power grid signal data demonstrates the framework’s ability to identify grid event signatures—distinctive patterns associated with faults and abnormal behaviors such as capacitor switching, load shedding, and inverter malfunctions, to name a few—while revealing potential connections between these events. In the case study, the performance of both TFT and VAE-based methods was evaluated under varying configurations, including latent vector dimensions (e.g., 8, 32, 64, 128, 256, and 640) and different dimensionality reduction algorithms. Two VAE-based methods were developed: one using 1D convolutional layers and the other leveraging LSTM for encoding multivariate time series data. Three validation methods are employed to assess the outputs of different models: internal validation metrics, a Human-in-the-Loop approach, and a relative validation technique that involves comparing outputs from various models using corresponding plots and shape-matching methods. The comparative analysis showed a consistency of 86–92\% between the 2D latent vector representations produced by TFT and the VAE-based methods under optimized model configurations and dimension reduction algorithms (e.g., t-SNE), and revealed that TFT outperforms VAE in run-time and scalability for diverse data shapes.  By integrating TFT with visual analytics, this work advances fault diagnosis in multivariate time series data, fostering explainable AI approaches that improve decision-making in critical system operations.


 

\section{Literature Review}
\label{Literature Review}
Analyzing patterns in multivariate time-series data, characterized by high dimensionality and complexity, requires interdisciplinary approaches to support effective decision-making. This literature review is structured into four sections. First, it identifies research needs for multivariate pattern analysis in urban and environmental studies. Second, it examines applications in power grid operations, where inherently multivariate data from diverse sensors provide practical use cases for disturbance diagnosis and pattern analysis. Third, it reviews AI and Human-in-the-Loop (HitL) methods for dimensionality reduction, enabling visual exploration and interpretation of complex datasets. Finally, it addresses challenges in current applications, emphasizing limitations in supporting experts with complex multivariate time-series data. Using power grid signal data as a case study, this review identifies knowledge gaps and drives advancements in fault diagnosis and multivariate pattern analysis.

\subsection{Multivariate Pattern Detection and Exploration in Broad Urban Management Applications}
Exploring multivariate patterns in urban and environmental data, often surrogates for underlying physical processes, is crucial for managing smart city systems. This approach supports hypothesis generation, anomaly detection, and fault diagnosis \citep{erhan2021smart, sarker2022smart}. By analyzing spatial and temporal variability across environmental attributes, infrastructure features, and urban characteristics, urban planners and environmental scientists can uncover insights into complex, interconnected systems such as transportation \citep{berres2021multiscale, pack2010visualization, berres2021explorative}, building energy \citep{xu2022geo, cottafava2018explorative}, public health \citep{xu2021episemblevis, preim2020survey}, and watershed management \citep{xu2022overview, xu2020web}. These insights drive the development of holistic management practices, enhancing the sustainability, resilience, and efficiency of urban environments \citep{bibri2021data}. 

To emphasize the importance of interpreting multivariate data patterns, we review notable studies in environmental and urban management. \citet{xu2019web} developed a web-based geovisual analytics platform leveraging machine learning and visual analytics to address culvert sedimentation, a key infrastructure issue in watershed management. The platform integrates diverse data sources and uses a tree-based feature selection algorithm and self-organizing maps (SOMs) to identify sedimentation drivers, visualize multivariate patterns, and inform mitigation strategies. This scalable, explainable AI tool is adaptable to broader environmental challenges. In the transportation sector, \citet{dadashova2021multivariate} used multivariate time series analysis to explore how socioeconomic factors impact urban traffic congestion. Using data from 51 U.S. metropolitan areas, they incorporated variables such as employment rates and fuel prices into vector autoregressive models with exogenous factors (VAR-X). These models provided interpretable insights for forecasting congestion metrics and informing targeted, data-driven management strategies. \citet{sanhudo2021multivariate} applied machine learning to improve building energy management through clustering and regression techniques. By using k-medoids clustering and dynamic time warping to enhance weather data quality, and Artificial Neural Networks (ANN) and Support Vector Regression (SVR) to rectify errors, the study demonstrated significant potential for energy optimization. However, limited explainability of these models restricts deeper insights into underlying patterns. Finally, \citet{hsieh2004nonlinear} extended classical techniques like PCA and CCA to nonlinear variants (NLPCA and NLCCA) using neural networks. Applied to climate and oceanographic data, these methods revealed intricate patterns in complex systems such as El Niño, offering valuable tools for managing multivariate data in smart grid systems and other resilient infrastructure applications.

Based on these example studies, it can be concluded that in various fields of urban and environmental studies, analyzing multivariate spatial and time-series data supports researchers and decision-makers in several key ways:
\begin{description}
    \item [Perform Dimensionality] reduction to group features based on multivariate similarities, facilitating the clustering of complex data. This approach aids in identifying environmental or urban phenomena by analyzing specific attributes of one or more data features \citep{xu2020web, sanhudo2021multivariate}. 
    \item [Explore Dependencies] among time series of multiple variables to enable causal inference, providing insights into the underlying physical processes represented by these data variables \citep{dadashova2021multivariate}.
\end{description}  
By analyzing surrogate data to understand environmental or urban phenomena and the interdependencies between physical processes, researchers can derive insights to inform effective management practices. This approach facilitates the promotion or mitigation of specific phenomena and their underlying processes, reducing potential hazards and faults in urban systems.

\subsection{Multivariate Pattern Detection and Exploration in Grid Operations}
As a critical urban subsystem, smart grids generate vast amounts of multivariate time-series data from various devices and sensors \citep{syed2020smart}. Smart meters at consumer sites record variables such as power consumption, voltage, and frequency, offering insights into energy usage patterns \citep{wang2020smart}. Digital Fault Recorders (DFRs) capture high-frequency recordings of disturbances, namely faults, on the order of microseconds, typically capturing point-on-wave (PoW) voltages and currents at various points in a distribution system. Phasor Measurement Units (PMUs) at substations typically stream measurements such as voltage, current, frequency, and phase angle, enabling real-time stability monitoring\citep{biswal2023real}. SCADA systems aggregate data from Remote Terminal Units (RTUs) and Intelligent Electronic Devices (IEDs) across substations, continuously monitoring grid health through variables like voltage, current, power flow, and equipment status \citep{manoj2021power}, though they report readings on the order of seconds rather then milliseconds, in the case of PMUs, or microseconds in the case of DFRs. Distributed Energy Resources (DERs), including solar and wind energy systems, provide data on generation levels, voltage, and frequency, which are critical for renewable energy integration. Environmental sensors add weather-related data—temperature, humidity, and wind speed—essential for predicting energy demand and supply \citep{nyangon2024climate}. Together, these data streams enable comprehensive grid analysis and control.

In the past decades, advanced time-series classification methods have been adopted into many power grids as a critical component of their monitoring and controls systems to facilitate anomaly detection, root cause analysis, and control purposes \citep{susto2018time}. 
For instance, \citet{ceci2020echad} introduces ECHAD (Embedding-based CHAnge Detection), an unsupervised machine learning method for detecting shifts in multivariate time-series data, such as energy metrics. ECHAD adapts dynamically to real-time patterns, improving resilience to collinearity and noise while monitoring renewable energy fluctuations and power loads. Although robust, its reliance on embeddings and one-class learning limits explainability. Similarly, \citet{pinzon2019real} proposes a Random Forest (RF)-based method for short-term voltage stability (STVS) assessment using PMU data. By classifying voltage states under disturbances like fast voltage collapses, the model provides interpretable alerts, aiding real-time decision-making. However, its reliance on predefined labels restricts its adaptability to new, unlabeled patterns.
\citet{alaca2024cnn} introduce a two-step phase fault classification method for analyzing power grid data using convolutional neural networks (CNNs). The first step involves fault detection, determining if a fault exists in the signal, while the second step identifies fault types, such as single-line-to-ground or line-to-line faults. Various feature extraction techniques, including amplitude and phase (AP), fast Fourier transform (FFT), wavelet transform (WT), and power spectral density (PSD), were explored to optimize performance. The study demonstrates that combinations like AP-AP and FFT-WT yield the highest accuracy. Simulation and real-world tests validate the method’s superiority over conventional one-step classification, highlighting its potential for real-time fault analysis in smart grid systems. 
\citep{bakdi2021real} introduces a novel data-driven methodology for analyzing multivariate time series data in grid-connected photovoltaic (GPV) systems. It employs Principal Component Analysis (PCA) for dimensionality reduction and decorrelation, followed by Kullback-Leibler Divergence (KLD) to detect anomalies. Kernel Density Estimation (KDE) refines the density estimation of transformed components (TCs) derived from PCA, enabling nonparametric and adaptive fault detection. This approach overcomes limitations of Gaussian assumptions, computational complexity, and parametric constraints. The methodology is validated through extensive experiments with real-world faults under Maximum and Intermediate Power Point Tracking (MPPT/IPPT) modes, demonstrating improved detection sensitivity, computational efficiency, and robustness to varying environmental conditions. Despite its high accuracy, both the methods are primarily designed to classify phase faults, rather than enabling an interactive ways to help domain experts explore and analyze different types of faults, and their potential connections.

Multivariate temporal pattern analysis in grid operation sectors often focuses on detecting data anomalies in continuous streams from devices such as Phasor Measurement Units (PMUs) and smart meters to extract operational insights and support decision-making. These analyses aim to enhance Fault Detection and Isolation (FDI), Predictive Maintenance, and Cybersecurity in grid management \citep{susto2018time}. Practical grid management faces several challenges that include heterogeneous data sources, nonstationary signals, computational complexity, scalability, and missing data. These challenges call for scalable, generative AI-powered methods capable of handling diverse grid time-series data efficiently to enable real-time, actionable insights for grid operations.

 
\subsection{Latent Space Cartography and Visual Analytics}
Deep generative models serve as universal tools for learning data distributions in high-dimensional spaces by mapping them to lower-dimensional latent spaces, enabling efficient representation and synthesis of complex data patterns \citep{frenzel2019latent}. To increase the explainability and interprebility of these deep generative models and their outputs, a few recent studies have provided methodologies to visualize the latent representations of the complex high-dimension data. The latent representation is a compressed, lower-dimensional encoding of input data that is produced by generative AI models captures its essential features and structures while discarding irrelevant details, it can carry valuable information to help researcher explains how the AI models proceeds the complex data and produce the analytical outputs, such as clusters and classifications. Methodologies presented through these studies are often known as the ``Latent Space Cartography'' and ``Latent Space Visual Analytics'' \citet{liu2019latent, kwon2023latent}, and have been developed to help users extract and analyze patterns in various types of data, emphasizing the interpretability and practical utility of latent spaces in diverse domains. 

As examples, \citet{frenzel2019latent} proposed metric-based transformations designed to improve the geometric structure of latent spaces generated by deep generative models, such as Variational Autoencoders (VAEs) and Generative Adversarial Networks (GANs). Their approach targets a significant gap in ensuring that latent space distances accurately represent semantic relationships, particularly in high-dimensional, non-Euclidean data like those encountered in natural language processing. This method enables enhanced clustering and interpolation, which are critical for applications such as language modeling and feature synthesis. \citet{liu2019latent} developed the concept of "Latent Space Cartography", an integrated visual analytics framework that facilitates tasks such as analogy construction, semantic dimension mapping, and latent space comparison. Designed for diverse datasets, including emojis, word embeddings, and biological features, their system employs user-defined semantic axes and advanced projection strategies, such as t-SNE and PCA, to uncover nuanced relationships within the data. By enabling interactive exploration, it bridges the gap between static visualizations and actionable insights, making it particularly valuable for scientific feature analysis and machine learning evaluation. \citet{kwon2023latent} introduced the "Latent Space Explorer", a multimodal visualization tool tailored for datasets such as cardiac MRIs and ECGs. This system provides a comprehensive workflow that integrates subgroup exploration, interactive decoding, and phenotype prediction accuracy analysis. By addressing the challenges of multimodal latent representation, the tool supports domain experts in uncovering subgroup-specific correlations and testing biases, thus enhancing clinical applications in cardiovascular health. \citet{o2022latent} further extended the latent space visualization paradigm by incorporating non-Euclidean geometries and multi-task metric learning to disentangle complex subgroup features in dynamic systems. Their methodology uses cartographic techniques to display latent spaces as two-dimensional visualizations, enabling users to examine patterns, correlations, and subgroup behaviors within fine-grained datasets. Applications include biological processes, where latent spaces reveal dynamic relationships affected by temporal or environmental factors, and physical systems, where domain knowledge informs geometric transformations for more robust interpretability.

These methodologies address key challenges in latent space analysis, such as enhancing interpretability, preserving geometric coherence, and accommodating specific data modalities. By emphasizing tailored visualization tools, they enable meaningful insights from complex data, driving advancements in AI-powered exploratory analysis across scientific, medical, and industrial fields.

\subsection{Knowledge Gaps and Motivation}
While existing methodologies in latent space cartography are valuable, they can benefit from further refinement. By integrating more advanced visual analytics techniques, such as sophisticated user interactions, enhanced analytical reasoning, and tailored visual encoding, these applications can be improved to handle more complex real-world urban data. Such enhancements would foster greater data interpretability and model explainability, supporting practical training, education, and decision-making in urban system management and operations. In the context of facilitating multivariate temporal pattern explorations to support piratical urban applications, such as the fault diagnosis in power grid data, the existing methods face the following limitations:
\begin{description} 
    \item [KG1. Lacking Efficient Methods for Multivariate Time-series Analysis:] Most existing studies employ GAN, VAE, and sentence transformer to analyze semantic data, text data, image data, and single variate time-series data. VAE and GAN are not proven to be computational efficient for analyzing complex multivariate time-series analysis 
    Studies that utilize dedicated 
    
    \item [KG2. Limited Scalability and Adaptibility to Varying Data Shapes:] The length and number of features in time-series data collected from various sensors across urban systems can vary significantly. Existing methods that rely on Variational Autoencoders (VAEs) and Generative Adversarial Networks (GANs) to derive latent vectors often lack scalability and generalizability for datasets with diverse shapes. These approaches frequently require substantial modifications to the AI models to adapt to differing dataset dimensions, posing challenges for efficient and flexible implementation.
    
    \item [KG3. Limited Visual Analytical Reasoning for Complex Data:] Most existing latent space cartography studies rely on simple 2D scatter plots with color coding to visualize the distribution of complex data after dimensionality reduction. However, this approach has significant limitations when it comes to representing data with multiple labels or diverse characteristics. The 2D color-coded point representation is inherently constrained, as it can visually encode only a very limited number of data features. Consequently, it fails to effectively capture and represent multi-faceted datasets, such as ensemble or multi-label data.
 
\end{description} 

Given these challenges and knowledge gaps, we are motivated to develop a visual analytical framework that leverages the Temporal Fusion Transformer (TFT)—a state-of-the-art generative AI model based on the transformer architecture for temporal data analytics—along with advanced visual analytics techniques. Our approach seeks to facilitate the extraction and exploration of multivariate temporal patterns in complex urban data. By harnessing the TFT's high efficiency, scalability, and adaptability for handling time-series data, we aim to address knowledge gaps KG 1 and KG 2. Through the design and experimentation of advanced visual analytics techniques, such as novel visual encodings and interactive user interfaces, we propose a comprehensive visual analytical workflow. This workflow will enable the effective visualization of complex latent vectors (embeddings) generated by the Temporal Fusion Transformer (TFT), including ensemble and multi-label data. By providing deeper insights into complex patterns, our proposed framework seeks to significantly enhance pattern exploration and interpretation in urban data systems, addressing knowledge gap KG 3. 

To evaluate the performance and accuracy of the TFT-based method, we implemented two VAE-based models—one utilizing a 1D convolution layer and the other employing an LSTM as the encoder—to validate the outputs generated by the TFT. Alongside this task, we developed an evaluation workflow that incorporates novel metrics and ensemble visualizations to comprehensively compare multiple aspects of the TFT and VAE-based methods.

\section{Methodology}
\label{sec: Methodology}

\subsection{Design Requirements}
\label{subsec:DesignRequire}
The target users of our framework include data scientists, domain experts, and management personnel across various urban systems. For instance, in the context of our case study using power grid data, grid operators responsible for managing urban energy systems would significantly benefit from this framework.

To address the identified challenges and knowledge gaps, we have established the following design requirements as the foundation of our approach:
\begin{description}
    \item [Scalability and Efficiency in Time-Series Analytics:]
    The proposed approach will leverage the Temporal Fusion Transformer (TFT) for its exceptional efficiency and scalability in processing multivariate time-series data. The method will ensure that the architecture adapts seamlessly to datasets of varying dimensions and complexities, providing a robust and versatile framework for diverse analytical scenarios.
    
    \item [Exploratory Visual Analytics:]
    Advanced visual analytics techniques will be integrated to enhance the exploration and interpretation of multivariate temporal patterns. The framework will feature advanced user interactions, data transformation processes, tailored visual encodings, and coordinated multiple views to effectively represent ensemble and multi-label data. This comprehensive approach aims to uncover intricate patterns and relationships within the data.

    \item [Interactive and Accessible User Interface:] Intuitive, interactive web-based tools, such as online visual dashboards, will be developed to facilitate user-driven exploration of temporal data patterns and latent vector embeddings. These tools will provide an accessible and user-friendly platform for dynamic data analysis, empowering users to uncover deeper insights and interact with complex datasets meaningfully and efficiently.

    \item [Enhanced and Customized Visual Encoding:]
    Innovative, customized visual encodings and representations will be designed to facilitate the visualization of multiple facets of the complex latent embeddings generated by the TFT. These visualizations will support the discovery of intricate temporal and feature-level patterns, addressing the limitations of traditional latent space cartography methods and enabling multi-faceted data analysis and interpretation.
    
    \item [Evaluation and Benchmarking:] Validation workflows will be implemented using generative AI models established in previous latent space cartography studies, such as VAE-based models, to benchmark the performance and accuracy of the TFT-based approach. Novel metrics and ensemble visualizations will be designed and employed to comprehensively compare temporal dependencies and feature extraction capabilities across models.
      
    \item [Multi-label Dependency Exploration:] The proposed methods will enable users to visually explore dependencies between multiple labels in multi-label time-series data by analyzing pattern similarities through distance metrics and other measures derived using the TFT. Real-world urban data often encompasses multiple labels created using taxonomies from different domains, which highlight interactions between diverse physical processes. For example, in power grid data, event signatures frequently involve multiple labels categorized based on various technical or scientific criteria, such as weather impacts, equipment conditions, and operational states. Capturing and visualizing these interdependencies will provide domain scientists with deeper insights and facilitate hypothesis generation.
    
\end{description}
These design requirements ensure that the proposed framework effectively addresses the challenges in extracting and interpret patterns from complex multivariate temporal data, while offering robust, scalable, and interpretable solutions for analyzing multivariate urban time-series.

\makebox[\textwidth][c]{
    \includegraphics[width=\textwidth]{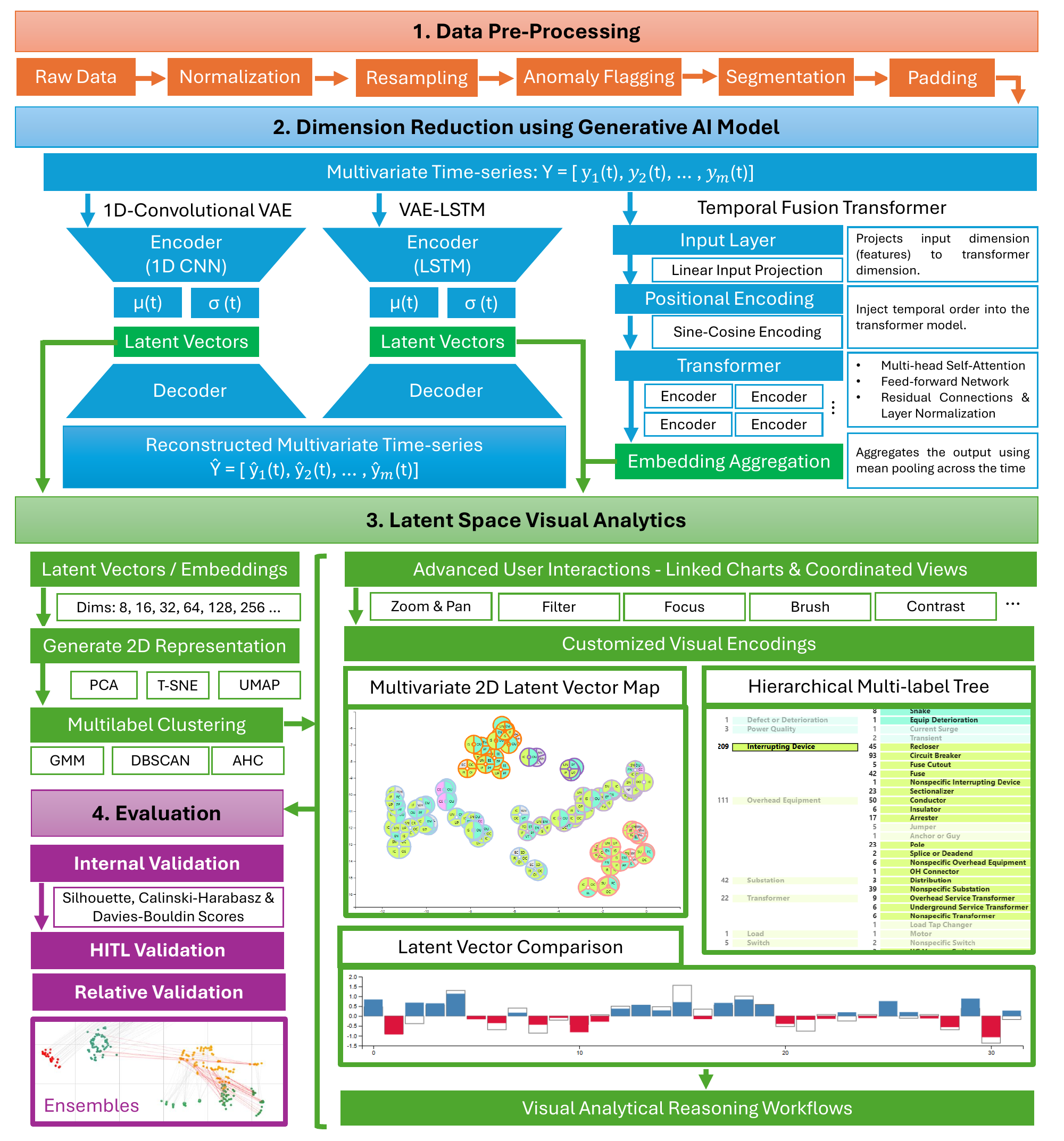}
}
\captionof{figure}{The conceptual design of our proposed framework.}
\label{fig:concepts_design}

\subsection{Framework Design}
\label{subsec:Data}
Our visual analytics framework comprises four key components: data preparation and preprocessing, generative AI models for dimensional reduction, visual analytical dashboards, and validation and evaluation metrics. The overall design of our framework is depicted in Figure \ref{fig:concepts_design}.  

\subsection{Data Preparation and Preprocessing}

Our framework is designed to analyze multivariate time-series data and can be adapted to facilitate pattern exploration across various disciplines. In this study, we demonstrate the framework's capabilities using power grid data as a showcase. The dataset originates from the Grid Event Signature Library (GESL), a resource developed by Oak Ridge National Laboratory (ORNL), Lawrence Livermore National Laboratory (LLNL), and Pacific Northwest National Laboratory (PNNL) under the U.S. Department of Energy's Office of Electricity. Additional details about this dataset are provided in Section \ref{subsec:uc-case-study}.

Similar to many compiled time-series datasets, the data collected from GESL consists of time-series measurements from diverse electrical systems provided by various data sources (e.g., utilities, industry, and academia). This diversity results in a heterogeneous structure with varying sampling rates, scales, and feature sets, posing challenges for consistent analysis and classification. All of the data contained in the GESL come from field measurements; no simulated data has been used in this study, or housed in the GESL.

To preprocess the data, we first address its heterogeneity by normalizing all numeric features to a uniform scale. Signals with varying sampling rates are resampled to a common rate—such as 20 ksps for certain providers—using linear interpolation, ensuring temporal consistency across the dataset. The time-series signals are then analyzed for anomalies by calculating differences between consecutive cycles. For example, using data from Provider 1, one cycle corresponds to 333 samples at 60 Hz. Deviations exceeding a predefined threshold are flagged as anomalies, enabling the identification of event-related irregularities.

To isolate baseline behavior, the data are segmented into 1-second intervals, ensuring uniform segment length. For files with insufficient data, padding is applied using the largest identified non-event region. Non-event segments are repeated as needed to achieve the required length, with smooth transitions ensured by aligning with the tail of the existing data. This preserves phase, amplitude, and temporal continuity. This segmentation and padding approach ensures all data segments are consistent and prepared for downstream analysis.

\subsection{Generative AI Models}
After preprocessing the data, we construct and apply two types of generative models with encoding capabilities: the Temporal Fusion Transformer (TFT) and a Variational Autoencoder (VAE). These models are used to perform dimensionality reduction, capturing multivariate temporal patterns within the time-series data with multiple features by deriving latent vectors across a range of dimensions. The optimal latent dimension is determined through heuristic exploration facilitated by a visual analytics interface. This interface enables users to interact directly with the generative AI models, experiment with different combinations of hyperparameters, and visually explain and interpret the results. In this study, we employed two types of VAEs: a 1D-Convolutional VAE and a VAE-LSTM, which are detailed through the following subsections. 

\subsubsection{TFT}
\label{subsec:method-component1}
The Temporal Fusion Transformer (TFT) is a specialized neural network architecture designed for temporal data representation and feature extraction \citep{lim2021temporal}. The model leverages a Transformer-based framework, starting with a linear input projection layer that maps input features into a high-dimensional embedding space optimized for sequence modeling \citep{cristian2024inter}. To preserve temporal order information, positional encodings are incorporated into the embeddings, enabling the model to effectively capture sequential dependencies. In recent studies, the TFT has been widely employed for prediction and forecasting tasks using multivariate time-series data \citep{lim2021temporal, liao2024tft}.

In this study, the core of the TFT architecture consists of Transformer encoder layers, which are utilized to model complex temporal dependencies and interactions across the sequence. These layers allow the model to effectively capture both short- and long-term relationships within the temporal data. To aggregate the temporal information, the output of the Transformer encoders undergoes mean pooling across the sequence length, producing a summarized representation. This representation can be further refined through an optional linear projection to generate the final embeddings for each input sample. 
Our TFT architecture is designed to process batches of temporal data with dimensions corresponding to the batch size, input feature dimension, and sequence length. It outputs latent embeddings that encapsulate both temporal and feature-level characteristics, making them suitable for a range of downstream tasks. In evaluation mode, the trained model computes embeddings for the input data, which are subsequently stored and mapped to corresponding filenames for further analysis. This architecture and workflow enable robust time-series representation learning, feature extraction for temporal prediction models, and domain-specific applications requiring detailed temporal embeddings. By employing Transformer encoders, the TFT demonstrates high expressivity and adaptability, making it particularly effective in capturing intricate temporal dependencies and providing a versatile solution for temporal data analytics.

\subsubsection{1D-Convolutional VAE}
\label{subsec:method-component2}
In this study, Variational Autoencoders (VAEs) are utilized to learn compact latent representations of multivariate time-series data, with an emphasis on capturing both temporal and feature-level dependencies. The 1D CNN VAE architecture consists of three main components: an encoder, a latent space representation, and a decoder. The encoder employs 1D convolutional layers to extract hierarchical temporal features from the input time series. These convolutional layers are followed by fully connected layers that parameterize the mean and log-variance  of the latent space distribution. The latent space is sampled using the reparameterization trick, which allows for differentiability and ensures stable optimization during backpropagation. The decoder reconstructs the original input data from the latent representations using transposed convolutional layers to upsample the data back to its original dimensions.

The training process is driven by a composite loss function that combines Mean Squared Error (MSE), which reconstruction loss, which measures the fidelity of the reconstructed output to the input data, and the Kullback-Leibler (KL) divergence. The KL divergence regularizes the latent space to follow a standard Gaussian distribution, enabling meaningful and compact representations.

The model is optimized over multiple epochs using the Adam optimizer with a low learning rate to ensure training stability. After training, the latent vectors are extracted for all samples, providing a compressed and semantically rich representation of the input time series. These latent vectors are subsequently saved and mapped to corresponding filenames for downstream applications. This implementation offers a robust framework for capturing and compressing the spatiotemporal structure of multivariate time series into a latent embedding space. These embeddings are highly versatile, supporting tasks such as anomaly detection, clustering, and predictive modeling, and demonstrating the effectiveness of the VAE for temporal data representation and feature extraction.

\subsubsection{VAE with a LSTM Layer}
\label{subsec:method-component3}
Building upon the VAE architecture with 1D convolutional layers described earlier, this research also explores a complementary approach by integrating a Variational Autoencoder (VAE) with an LSTM-based architecture. This variant is designed to handle temporal data with a focus on preserving sequential dependencies, making it particularly suited for time-series data with intricate temporal dynamics.

The encoder employs bidirectional LSTM layers to capture both forward and backward temporal relationships in the data. This design ensures a comprehensive understanding of sequential dependencies across the entire time horizon. The sequential output from the LSTM layers is then flattened and passed through a fully connected layer, which parameterizes the latent space by computing the mean and log-variance of the latent distribution. The latent space representation is sampled using the reparameterization trick, facilitating smooth gradient flow during backpropagation while maintaining differentiability. The decoder reconstructs the input sequences from the latent vectors using a combination of fully connected and LSTM layers. This reconstruction pathway is carefully designed to preserve the temporal structure of the original data, ensuring accurate sequence generation.

Similar to the VAE with 1D convolutional layers, the training process optimizes a composite loss function that balances MSE and KL divergence. The model is trained over multiple epochs using the Adam optimizer, with batched data to enhance computational efficiency. After training, the latent vectors are extracted, serving as compressed representations of the input sequences for downstream tasks such as clustering, anomaly detection, or predictive modeling. In addition to handling temporal data, this implementation incorporates metadata processing to enhance analysis. Event tags associated with the input sequences are parsed to generate binary vectors indicating the presence of specific events. These vectors are visually encoded using smooth color gradients, improving interpretability and facilitating the exploration of relationships between latent vectors and metadata. A mapping from filenames to event tags, binary arrays, and associated properties supports detailed metadata analysis, integrating advanced deep learning with visual analytics.

This VAE-LSTM approach complements the VAE with 1D convolutional layers by offering an alternative architecture that excels in tasks requiring explicit modeling of long-term sequential dependencies, thereby enriching the analytical capabilities for temporal data representation and interpretation. Both of them are later used as the benchmark to evaluate the performance of the TFT. 

\begin{figure*}[htb]
 \centering
\includegraphics[width=\textwidth]{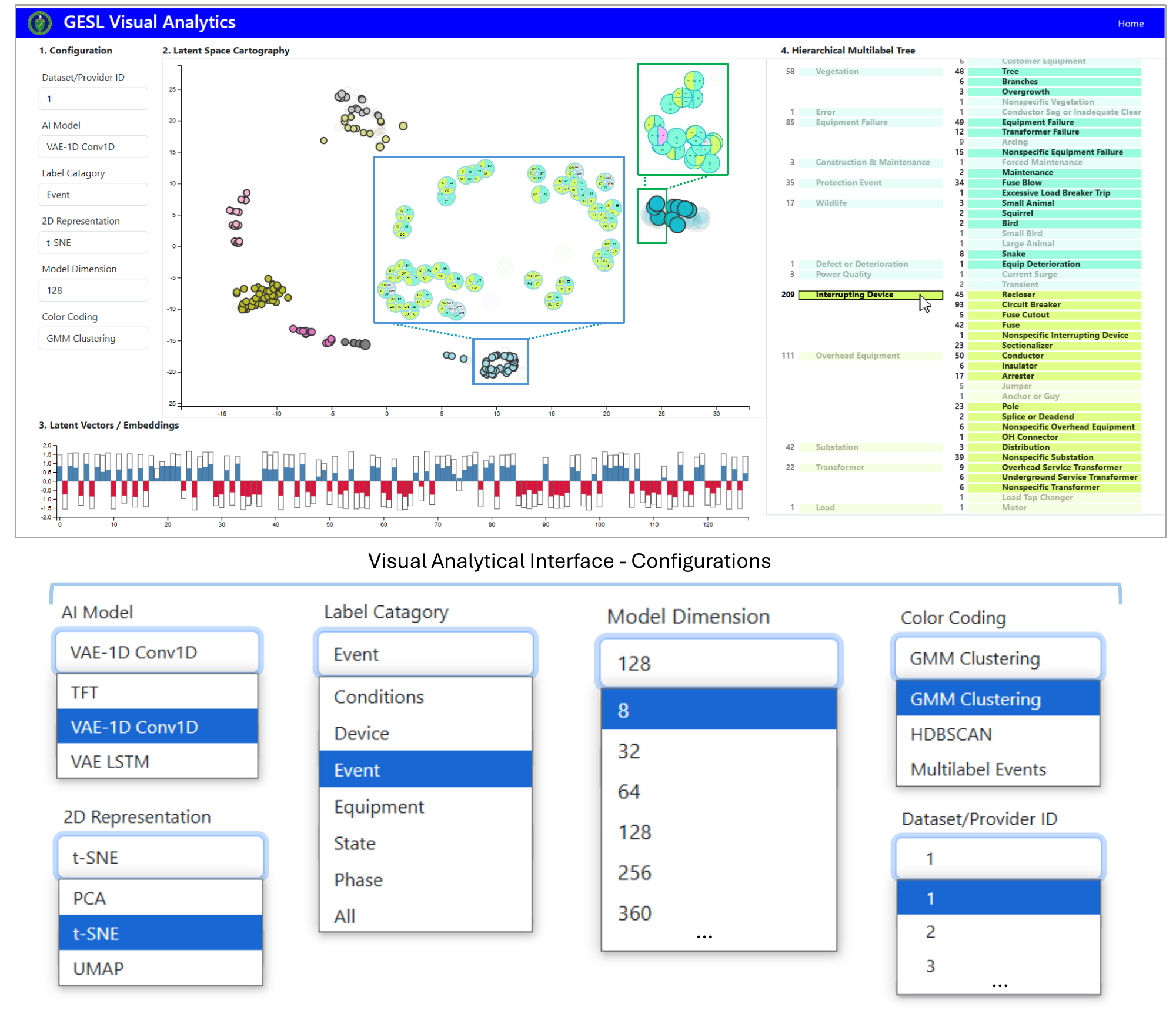}
 \caption{The overview of the visual analytics interface for exploring multivariate temporal patterns in multi-label power grid data. }
 \label{fig:interface-overview}
\end{figure*}

\subsection{Latent Space Visual Analytics}
\label{subsec:method-component4}
A visual analytics interface with linked charts and multiple coordinated views has been developed to enable users to visually explore and analyze the latent vectors (embeddings) generated by the TFT and VAEs. An overview of the interface is shown in Figure \ref{fig:interface-overview}. 

Within the interface, we employed customized visual encoding to accomodate the analytical needs and the characteristics of the complex data. Visual encoding, also known as visual representation, refers to the use of graphical elements such as position, color, size, and shape to represent data in a way that enables intuitive exploration and analysis. Given the multilabel nature and complexity of the multivariate time-series data analyzed in our framework, we devised two novel visual representations to enhance existing visualization techniques commonly used in latent space cartography applications. These innovative visual encodings include the Multivariate 2D Latent Vector Map and the Hierarchical Multi-label Tree.

\makebox[\textwidth][c]{
    \includegraphics[width=\textwidth]{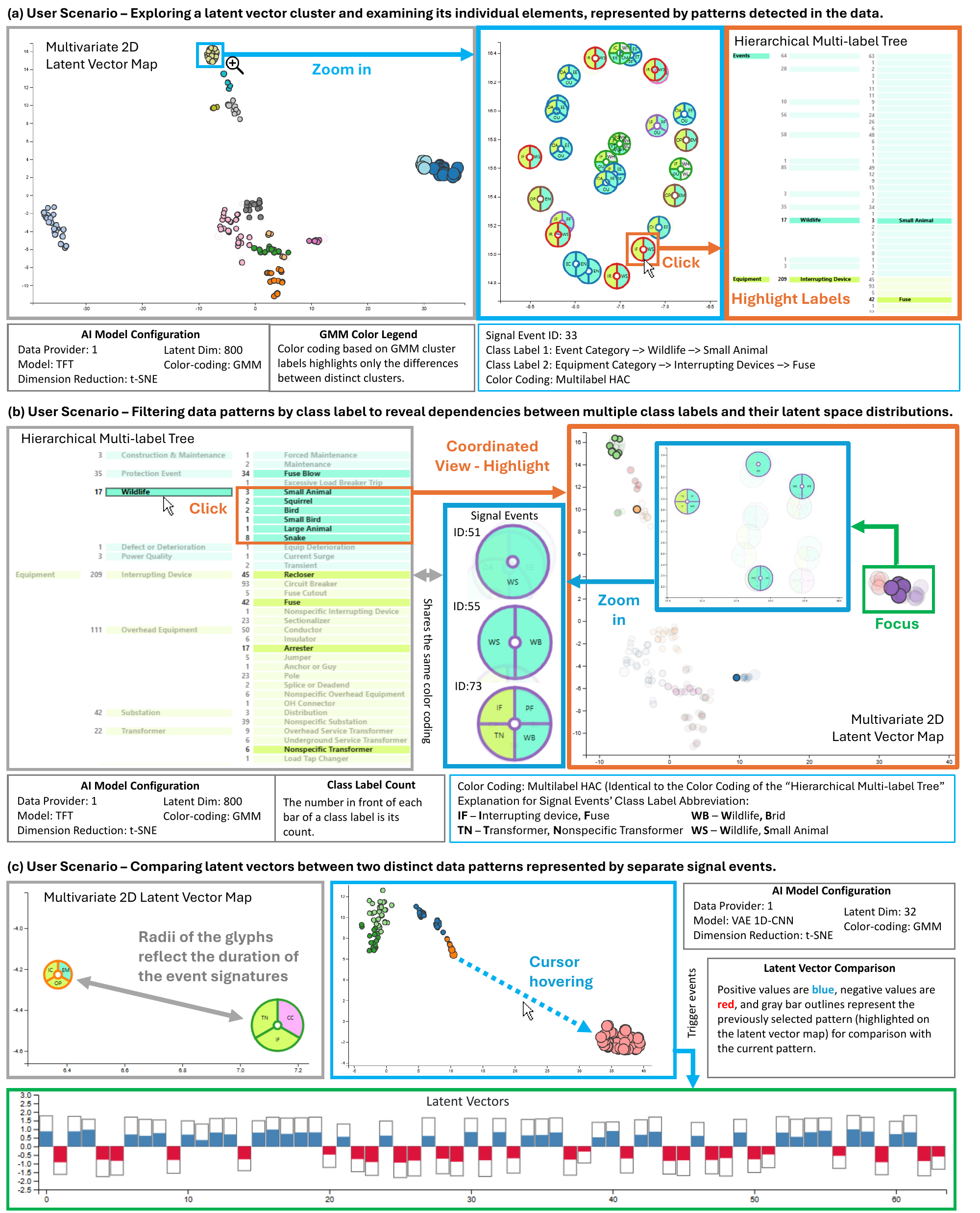}
}
\captionof{figure}{Three scenarios are presented to demonstrate the design and user interaction features of the customized visual encoding developed for the proposed framework.}
\label{fig:visual encoding comparasion}

\subsubsection{Multivariate 2D Latent Vector Map}
The Multivariate 2D Latent Vector Map provides a spatial representation of the latent vectors (embedding) derived from VAE or TFT. High-dimensional latent vectors are projected into a 2D space as latent representation using dimensionality reduction techniques. These techniques include PCA, t-SNE, or UMAP, which can be selected via the visual analytical interface. The visualization features an interactive viewport, resembling web-based map engines like Leaflet, which allows users to explore clusters, outliers, and temporal transitions within the data through interactions such as zooming, panning, and focusing on specific clusters and their elements (2D latent representtion displayed on the map), as depicted in Figure \ref{fig:visual encoding comparasion}a. This representation offers valuable insights into the temporal and feature-level dependencies captured by the models. To prevent visual clutter and ensure users receive the appropriate level of detail, the map dynamically toggles between different visual representations based on the zoom level. At a low zoom level, which provides a global overview of the latent vector distribution, the visualization uses color-coded dots to represent data points. The color-coding options include categorical cluster labels generated by clustering algorithms such as Gaussian Mixture Models (GMM), Density-Based Spatial Clustering of Applications with Noise (DBSCAN), and agglomerative hierarchical clustering (AHC) applied to class labels. These color-coded dots effectively highlight broad patterns and relationships within the data. At higher zoom levels, where users focus on specific regions to examine local latent vector distributions, the visualization employs glyph-based representations (depicted in the blue box in the user scenario b in Figure \ref{fig:visual encoding comparasion}). These glyphs encode multiple class label attributes into the 2D latent vectors, allowing users to explore the relationships, dependencies, and correlations among complex data patterns. The radius of the dot and the glyph is used to encode the duration of the signal event underlying each 2D latent representation. 

This enhanced local view aids in identifying potential connections or inconsistencies across multi-label data, fostering a deeper understanding of intricate data patterns and relationships.

\subsubsection{Hierarchical Multi-label Tree}
The Hierarchical Multi-label Tree encodes the structure of class labels from the multi-label time-series data into a hierarchical tree representation using color-coded bars. Inspired by dendrogram visualizations, this design aims to reveal the hierarchical relationships among class labels while applying a hierarchical color-coding scheme to represent individual class labels according to their respective categories. This visualization helps users understand the relationships between labels, their hierarchical organization, and the underlying patterns associated with different data segments.

Through advanced user interaction techniques, users can select specific bars representing class labels using their cursor. The visualization then highlights all other class labels that co-occur with the selected label, providing an intuitive exploration of label dependencies. Simultaneously, through coordinated views, the Multivariate 2D Latent Vector Map highlights the 2D distribution of latent representations corresponding to data patterns containing the selected class label. This interaction is illustrated in the user scenario (b) in Figure \ref{fig:visual encoding comparasion}. Since the Hierarchical Multi-label Tree is linked to the Multivariate 2D Latent Vector Map, selecting a 2D latent representation on the map will automatically highlight its associated class label in the tree. This bidirectional interaction is demonstrated in the user scenario (a) in Figure \ref{fig:visual encoding comparasion}. By seamlessly linking hierarchical and spatial representations, the tool enables users to drill down into specific subsets of the data while maintaining a contextual overview of the entire dataset, fostering both granular analysis and holistic understanding.


\subsubsection{Latent Vector Comparison}
To enhance the explainability of the generative AI models, we incorporated an enhanced bar chart into the visual analytics interface. This bar chart allows users to visually compare the latent vectors of the original latent dimensions across different signal event patterns, which are represented as 2D latent projections in the Multivariate 2D Latent Vector Map. The bar chart is also dynamically linked to the Multivariate 2D Latent Vector Map and visualizes the latent vectors of the last two 2D latent representations selected by the user.

The latent vectors of the currently selected 2D latent representation are color-coded, with blue representing positive values and red representing negative values, contrasted against a gray outline that visualizes the latent vector values of the previously selected 2D latent representation. While the absolute values of the latent vectors do not carry inherent physical meaning, the similarity and variability of these vectors—whether across different clusters or within the same cluster—can provide insights into how the TFT or VAE-based model performs dimensional reduction on multivariate time-series data.

Additionally, users can export the latent vectors as files, enabling further analysis to improve the interpretability and interoperability of the AI models. This feature not only facilitates a deeper understanding of the models' decision-making processes but also supports advanced analytics for cross-disciplinary applications.


\subsection{Validation and Evaluation Metrics}
\label{subsec:method-component5}
To evaluate the quality of the clusters formed from the latent space vectors generated by the TFT, we developed a comprehensive validation framework comprising three strategies: Internal Validation using quantitative metrics, Human-in-the-Loop Validation leveraging visual analytics and domain expertise, and Relative Validation through comparative analysis of results produced by the two approaches.

\subsubsection{Internal Validation}
Internal Validation focuses on assessing the quality of clusters using information inherent to the dataset, without reference to external ground truth labels. we employ three widely used clustering evaluation metrics: Silhouette Score, Calinski-Harabasz Score, and Davies-Bouldin Score. These clusters are derived by further dimensional reduction of the latent space vectors using methods such as Principal Component Analysis (PCA), Uniform Manifold Approximation and Projection (UMAP), and t-Distributed Stochastic Neighbor Embedding (t-SNE). The Silhouette Score measures how well-separated the clusters are by comparing intra-cluster cohesion to inter-cluster separation, with higher values indicating better-defined clusters. The Calinski-Harabasz Score evaluates the ratio of between-cluster dispersion to within-cluster dispersion, rewarding well-separated clusters with low internal variance. The Davies-Bouldin Score assesses the average similarity ratio of each cluster to its most similar cluster, with lower values reflecting more compact and distinct clusters. These metrics collectively provide a robust framework for quantitatively assessing the quality of clusters in the reduced-dimensional latent space, enabling the validation of the effectiveness of TFT and VAE in capturing meaningful structures in multivariate time-series data.

\subsubsection{Human-in-the-loop Validation}
Human-in-the-Loop Validation integrates visual analytics and domain expertise to evaluate the quality and relevance of clusters formed in the latent space. This approach complements the quantitative metrics used in internal validation by leveraging human insight and justification to identify patterns, anomalies, and domain-specific structures that may not be accurately captured by AI models.

To facilitate this process, we employ advanced visualization techniques such as scatter plots enhanced with tailored multivariate visualization features. These enable users to explore and examine the reduced-dimensional representations of latent vectors and embeddings generated by TFT and VAE-based approaches. Dimensionality reduction is performed using methods such as PCA, UMAP, and t-SNE, and is augmented with advanced user interaction capabilities. These features allow users to pan, zoom, and focus on specific regions of the latent vector map through an interactive viewport. Users can further refine their analysis by hovering over individual data points, each representing a latent vector of the multivariate time-series data. This action dynamically triggers the coordinated display of linked charts and views, revealing detailed information such as corresponding labels and associated metadata. These interactive features provide an intuitive and comprehensive framework for users to analyze and interpret the underlying structure of the latent space, enabling deeper insights into complex data.

The visualizations are implemented through a web-based visual dashboard, allowing domain experts to qualitatively assess cluster separability, density, coherence, and the consistency and interrelationships of class labels within each cluster. For instance, in the context of fault diagnosis, domain experts such as grid operators can identify meaningful patterns or outliers within the latent vector representations, uncovering connections to real-world phenomena. One example identified through analyses using our framework is the consistent clustering of grid events involving arcing—a phenomenon of electrical current flow—and jumper failure, which refers to the failure of flexible or rigid conductive cables or connectors. These data-driven insights suggest a potential physical interaction between the two phenomena. This hypothesis can be validated using domain knowledge from the power grid, thereby confirming the AI model’s outputs and ensuring alignment with real-world behavior. The insights gained through this iterative, human-centered validation process are instrumental in refining the model and its outputs. By incorporating expert expertise, Human-in-the-Loop Validation ensures that clustering results are not only statistically robust but also contextually meaningful, effectively bridging the gap between computational analysis and domain-specific understanding. This example is later detailed as one of the use cases in Section \ref{subsec:uc-arching-jumper}.


\subsubsection{Relative Validation}
\label{subsec:Relative Validation}
To ensure the reliability and trustworthiness of the clustering results derived from the TFT and VAE-based methods, we developed an interactive visual analytical interface that enables domain experts to validate and interpret the results. This interface facilitates the integration of class labels with domain knowledge to provide meaningful justifications for the clustering outcomes. Additionally, we employed a relative validation approach to assess the consistency and similarity of the clustering outputs generated by the two methods. This approach utilizes quantitative cluster similarity metrics alongside a specialized visualization technique, known as the correspondence plot, which highlights the alignment and differences between the clustering results. By connecting corresponding points across the two methods, this visualization provides an intuitive way to compare the distributions and structural relationships of clusters, thereby enabling a robust evaluation and enhanced interpretability of clustering performance. The methodological details of our approach are as follows:
\begin{description} 
\item[Cluster Similarity Metrics] is designed to quantify the similarity between latent vector clusters derived from TFT and VAE by comparing the neighbors of individual latent vectors within their respective clusters. Neighbors are identified using the K-Nearest Neighbors (KNN) algorithm, applied to the 2D distributions of latent vectors after dimensionality reduction using methods such as PCA, t-SNE, and UMAP. Cluster labels are generated using the DBSCAN algorithm, based on the 2D latent vector distributions. A constraint is applied to ensure that only latent vectors with the same cluster label are considered neighbors. The similarity metric is then calculated as a percentage, reflecting the proportion of identical neighbors for each latent vector across the TFT and VAE-based approaches. This approach provides a quantitative measure to assess the consistency and alignment of cluster structures between the TFT and VAE-based methods. 
\item[Correspondence Plot] is the visualization technique designed to compare the distribution and clustering of two sets of point clouds by visually mapping corresponding points across the datasets. In our study, we use the correspondence plot to analyze and compare the latent vectors produced by the VAE- and TFT-based approaches. The plot not only highlights the spatial distribution and clustering of points derived from each method but also employs Dynamic Time Warping (DTW)-styled lines to connect corresponding points—latent vectors with the same ID but derived using the two different methods. These DTW-styled lines facilitate shape matching by capturing the similarity or divergence between the geometric shapes and distributions of the two point clouds. By analyzing the directional alignment and distance of these lines, we can statistically assess the deformations and shifts in clustering patterns. This allows us to compare the structural consistency of the clustering outputs produced by the VAE and TFT-based approaches, providing insights into the reliability and robustness of the methods.
\end{description}
The validation results from our case studies are detailed in Subsection \ref{subsec:valid-relative}, where the outputs of TFT and VAE-based methods are systematically compared across various configurations, including different latent dimensions and dimensionality reduction algorithms.

\section{Result Discussion}
\label{subsec:Result}
We applied our framework to assist energy scientists and power grid operators in visually exploring patterns within grid data to facilitate fault diagnosis. The use cases are designed to demonstrate the usability and practical benefits of our approach, which leverages visual analytics to place humans in the loop. This approach enables users to supervise and interact with AI applications that increasingly automate data analytics, ensuring greater reliability, interpretability, and decision-making support.

\subsection{Case Study Background}
\label{subsec:uc-case-study}
The Grid Event Signature Library (GESL) is an open-access repository designed to provide comprehensive, high-resolution measurement data for power system events. Sponsored by the U.S. Department of Energy, GESL includes over 5,600 event records, with labeled and unlabeled data from diverse sources, including Phasor Measurement Units (PMUs) and Point-on-Wave (PoW) devices. These datasets encompass a variety of grid behaviors, including transient events like capacitor switching, load shedding, and frequency oscillations, along with ambient operational conditions. 

The GESL incorporates several key features designed to enhance its utility for research and practical applications. Each event record is enriched with detailed metadata, including sampling rates, event durations, and textual descriptions, ensuring comprehensive contextual information. The repository integrates diverse data types from multiple measurement technologies, such as Phasor Measurement Units (PMUs), Power-over-Wire (PoW) sensors, and specialized GridSweep™ devices, enabling the capture of both sinusoidal and non-sinusoidal grid behaviors. A hierarchical taxonomy is employed to categorize events based on physical and operational phenomena, providing structured labeling that enhances the interpretability and consistency of the datasets. Furthermore, GESL supports the development of AI-driven applications by offering open access to high-quality data, facilitating analysis, training, and benchmarking for researchers and practitioners alike.

The significance of GESL lies in its ability to bridge the gap between operational data and research needs. By offering labeled and unlabeled datasets, it supports applications in event detection, clustering, anomaly analysis, and real-time grid monitoring. Its availability and diversity make it a vital resource for advancing data-driven methodologies in power system analytics and the integration of renewable energy sources.

\subsection{Use Cases}
We pre-processed and analyzed multiple datasets collected from six GESL providers, and created two use cases to demonstrate the practical value of our proposed method. 

\subsubsection{Inter-pattern Dependency Analysis}
\label{subsec:uc-arching-jumper}
Our proposed visual analytics framework enables users to explore potential interconnections and dependencies between power grid fault event patterns through linked charts and coordinated views (illustrated in Figure \ref{fig:usercase-1}). The following use case illustrates how a user can interact with the system to analyze complex relationships within the fault event data. The case study is developed using the data and its augmented version from GSEL Provider 1, which includes six data features: three current fields ('Current.Ia', 'Current.Ib', and 'Current.Ic') and three voltage fields ('Voltage.Va', 'Voltage.Vb', and 'Voltage.Vc'). The dataset also includes a timestamp column, 'Time µs', which provides the temporal context for the time-series data. These six features form the multivariate time-series data used for analysis in the study.

\makebox[\textwidth][c]{    \includegraphics[width=\textwidth]{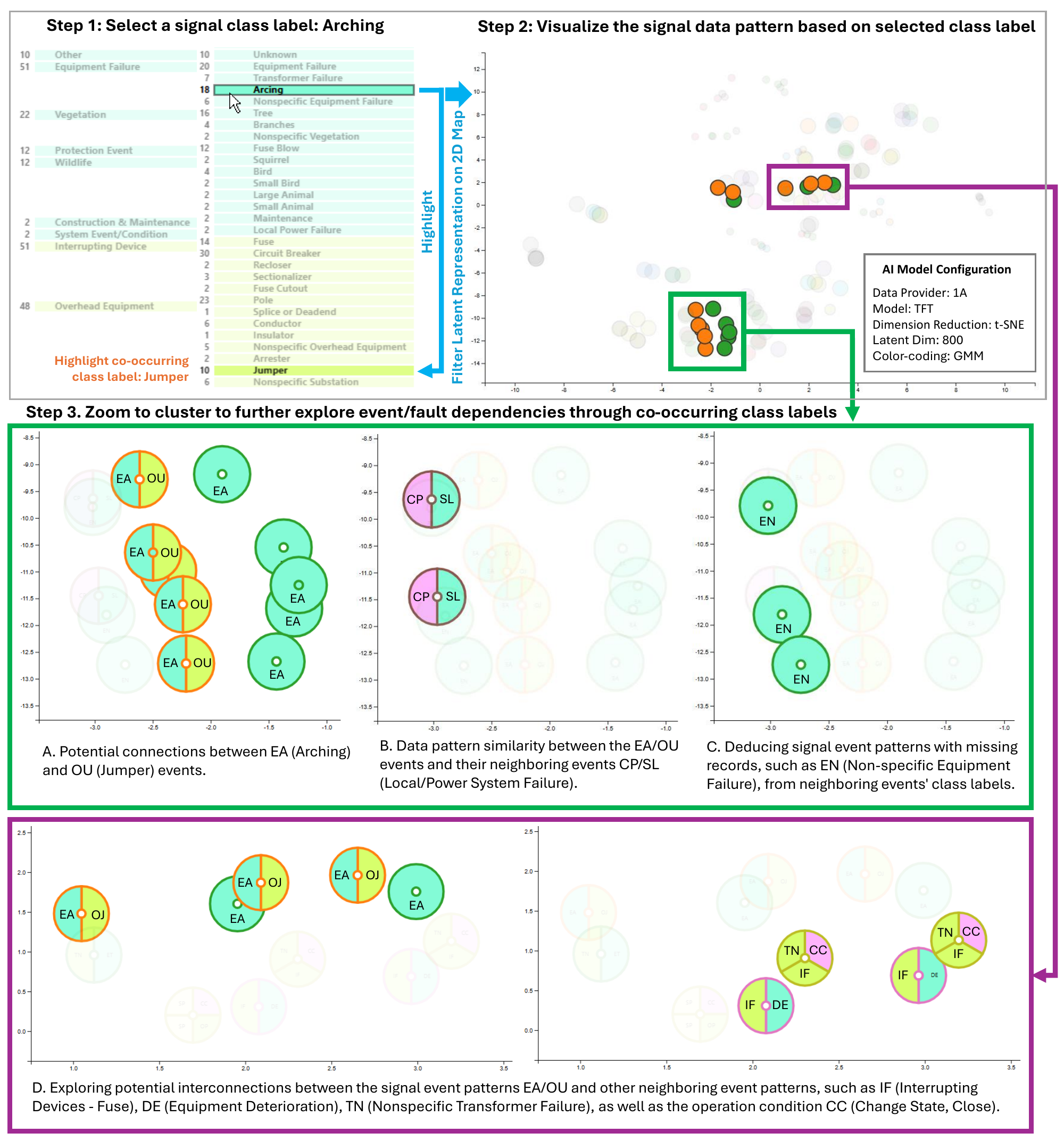}
}
\captionof{figure}{Use Case 1 and 2: Deducing potential interconnections and dependencies between different fault event patterns, and facilitate the imputation of missing labels for certain fault event patterns.}
\label{fig:usercase-1}

\begin{description}
    \item[Step 1: Selecting a Class Label of Interest:] The user begins by interacting with the Hierarchical Multi-label Tree to select a class label of interest. In this scenario, the user selects the class label "Arching". Upon selection, the visualization automatically highlights the co-occurring class label "Jumper", which frequently appears alongside the selected class label in the fault event data. This linked interaction provides an immediate understanding of class label relationships.
    \item [Step 2: Visualizing Latent Representations:] Following the class label selection, the Multivariate 2D Latent Vector Map automatically highlights the corresponding 2D latent representations of the fault events. These representations are derived by reducing the dimensionality of the multivariate time-series data using the Temporal Fusion Transformer (TFT) model. In this case, the user selects t-SNE as the dimensionality reduction method to project the high-dimensional TFT latent vectors (e.g., 640 dimensions) into a 2D space. This projection allows users to observe the spatial distribution and relationships of the latent representations, offering insights into potential dependencies and patterns.
    \item [Step 3: Exploring Spatial Relationships and Dependencies] The user zooms into the clusters on the 2D latent vector map to examine their spatial organization and identify neighboring clusters with different class labels. This spatial proximity may reveal potential interconnections and dependencies between fault event patterns. In this scenario, the latent vectors for the "Arching" and "Jumper" fault events are observed to be spatially close to neighboring clusters representing other class labels, such as "Local/Power System Failure" (CP/SL), "Interrupting Devices - Fuse" (IF), "Non-specific Transformer Failure" (TN), and "Event Deterioration" (DE).
    \item [Step 4: Deduction] The spatial proximity of these clusters suggests similarities or correlations in the underlying fault patterns, which may indicate shared causes or interdependent events. Such insights can be further interpreted and explained using power grid domain knowledge to uncover relationships between different fault types, such as equipment failures, system interruptions, or signal deterioration.
\end{description}

This use case highlights the effectiveness of our framework in integrating generative AI models with interactive visual analytics. By linking the Hierarchical Multi-label Tree with the Multivariate 2D Latent Vector Map, the system allows users to dynamically explore relationships between fault event patterns. The framework not only facilitates the identification of co-occurring fault events but also supports the discovery of hidden dependencies and anomalies within the data. This human-in-the-loop approach ensures that power grid operators and energy scientists can make informed decisions, improving fault diagnosis and grid reliability. The domain knowledge justifications for the data-driven dependencies between various fault class labels are further elaborated in Section \ref{subsec:hitl-validation}.

\subsubsection{Imputation of Missing Fault Class Labels }
\label{subsec:uc-classification}

Our framework demonstrates significant value in supporting the imputation of missing fault class labels by leveraging its ability to deduce signal event patterns with incomplete records, such as EN (Non-specific Equipment Failure), from neighboring events' class labels. This user scenario is demonstrated in task D in Figure \ref{fig:usercase-1}. By analyzing the spatial proximity of latent representations in the Multivariate 2D Latent Vector Map, the framework identifies relationships between events with missing or ambiguous labels and adjacent well-defined class labels, such as EA (Arching), OU (Jumper Failure), or DE (Equipment Deterioration). This approach allows users to infer the most likely fault class for data points lacking annotations, based on shared latent features and clustering behavior in the reduced-dimensional space. By incorporating domain knowledge and these derived dependencies, the framework facilitates accurate label imputation, enhancing the completeness and reliability of fault datasets. This capability is particularly valuable for training AI models, improving the interpretability of power grid fault analyses, and addressing data gaps in real-world operational contexts.

With the ability to visualize the latent vectors of data patterns with missing labels and compare them to the 2D latent representations of neighboring data patterns, our proposed visual analytics methods provide a more explainable approach for facilitating missing label imputation. By incorporating human expertise into the loop through latent vector visualization, our approach enhances the interpretability of AI models, fostering greater trust in the results and significantly improving the reliability of the imputation process.

\subsubsection{Model Parameter Tuning}
\label{subsec:uc-ai_tuning}
The proposed visual analytics methods provide a powerful interface for visually comparing and assessing the quality of clusters generated from the TFT's dimensional reduction of multivariate time-series data. By allowing users to explore the results under different configurations, such as varying the latent dimensions or the dimensional reduction methods used to project high-dimensional latent vectors into 2D representations, the interface offers a flexible and intuitive way to evaluate model performance. Users can dynamically adjust configurations through the interface and visually analyze how these changes impact the clustering results. 

As demonstrated in our use case with signal event data from Provider 4, we compared clusters produced by TFT combined with three dimensional reduction methods that include PCA, t-SNE, and UMAP, under two latent dimensions (8 and 256). This comparative analysis revealed distinct differences in cluster quality and separation, enabling a heuristic approach to identify the most effective configurations for dimensional reduction and latent space representation (depiced in Figure \ref{fig:usercase-3}). 
Through the visualization, users can identify model configurations that achieve clear separation of signal events with different class labels. For instance, points color-coded in blue represent events labeled as SG (Source, Generator) and CT (Change State, Trip), while points color-coded in orange correspond to events labeled as CS (Change State, Trip). Datasets from Provider 4 and several other providers, which are well-labeled and thoroughly documented, can be used as benchmark datasets to evaluate AI model performance and optimize parameter configurations. In this use case, specific combinations, such as "t-SNE with a latent dimension of 8", can be identified as optimal setups for clustering tasks within our framework, as they produce clusters with minimal overlap between events with different labels. 

While the use case focuses on a specific dataset and parameter set, the framework is versatile and supports broader exploration across diverse latent dimensions and configurations. For datasets with missing labels, such as those from Provider 1, users can subset the data with well-recorded event class labels and use this subset to tune model parameters, optimizing clustering performance. Once the model is tuned, the full dataset can be introduced for additional insights and heuristic exploration, enabling users to uncover hidden patterns and improve the interpretability of incomplete datasets. This iterative approach ensures robust analysis even with partially labeled data.

This capability empowers users to iteratively refine AI model parameters, enhancing their ability to identify optimal setups for clustering and improving overall model interpretability and performance. In addition to the visual analytics approach, our framework incorporates quantified metrics through internal validation techniques to assess and characterize the clustering results produced by both the VAE and TFT models. These internal validation methods provide a complementary, data-driven evaluation of cluster quality, enhancing the robustness of the analysis. Details of these validation techniques are presented in Section \ref{subsec:internal-validation}.

\subsection{Validation and Performance Comparison}
Through the case study, we utilize data from GSEL Data Provider 1 to showcase the validation process within our framework. This dataset is particularly well-suited for this purpose as it contains a rich diversity of class labels, enabling a comprehensive demonstration of the framework's ability to handle complex, multivariate time-series data and validate clustering performance of the TFT and VAE models effectively.

\subsubsection{Internal Validation}
\label{subsec:internal-validation}
The validation process evaluates clustering results produced by TFT, VAE-1D CNN, and VAE-LSTM models under varying configurations of latent dimensions and dimensionality reduction algorithms (PCA, t-SNE, and UMAP) using internal validation metrics. These metrics assess cluster quality quantitatively, focusing on metrics like compactness, separation, and silhouette scores to characterize the effectiveness of the clustering. By comparing the clustering performance across different configurations, the process identifies optimal parameter combinations for generating well-separated and interpretable clusters, ensuring robust model evaluation and enabling users to fine-tune AI models for improved performance.

\subsubsection{Human-in-the-loop Validation}
\label{subsec:hitl-validation}
Using the use case presented in Section \ref{subsec:uc-arching-jumper} as an example, the spatial proximity of the 2D latent representations for "Arching" and "Jumper Failure" can be effectively explained and interpreted through the lens of power grid management domain knowledge. In power grid operation and urban power system management, "Arching" and "Jumper Failure" are closely interrelated due to shared physical processes and operational dependencies. Jumper failures, which involve the breakdown or disconnection of flexible conductors, often create unstable circuits, resulting in abnormal voltage conditions that trigger arcing—electrical discharges across exposed or loose connections. Conversely, the intense heat generated by arcing can degrade insulation and weaken nearby jumpers, leading to further failures. These events frequently co-occur in both temporal and spatial proximity, particularly at points of mechanical stress, aging infrastructure, or under transient conditions such as switching surges or lightning strikes. Environmental factors, such as thermal cycling, wind loads, and vibrations, exacerbate these failures, particularly in older urban systems. Understanding the dependency between these fault patterns enables grid operators to improve fault diagnosis, identify root causes, and implement predictive maintenance strategies, ultimately enhancing the resilience and reliability of power distribution systems.

\makebox[\textwidth][c]{
    \includegraphics[width=\textwidth]{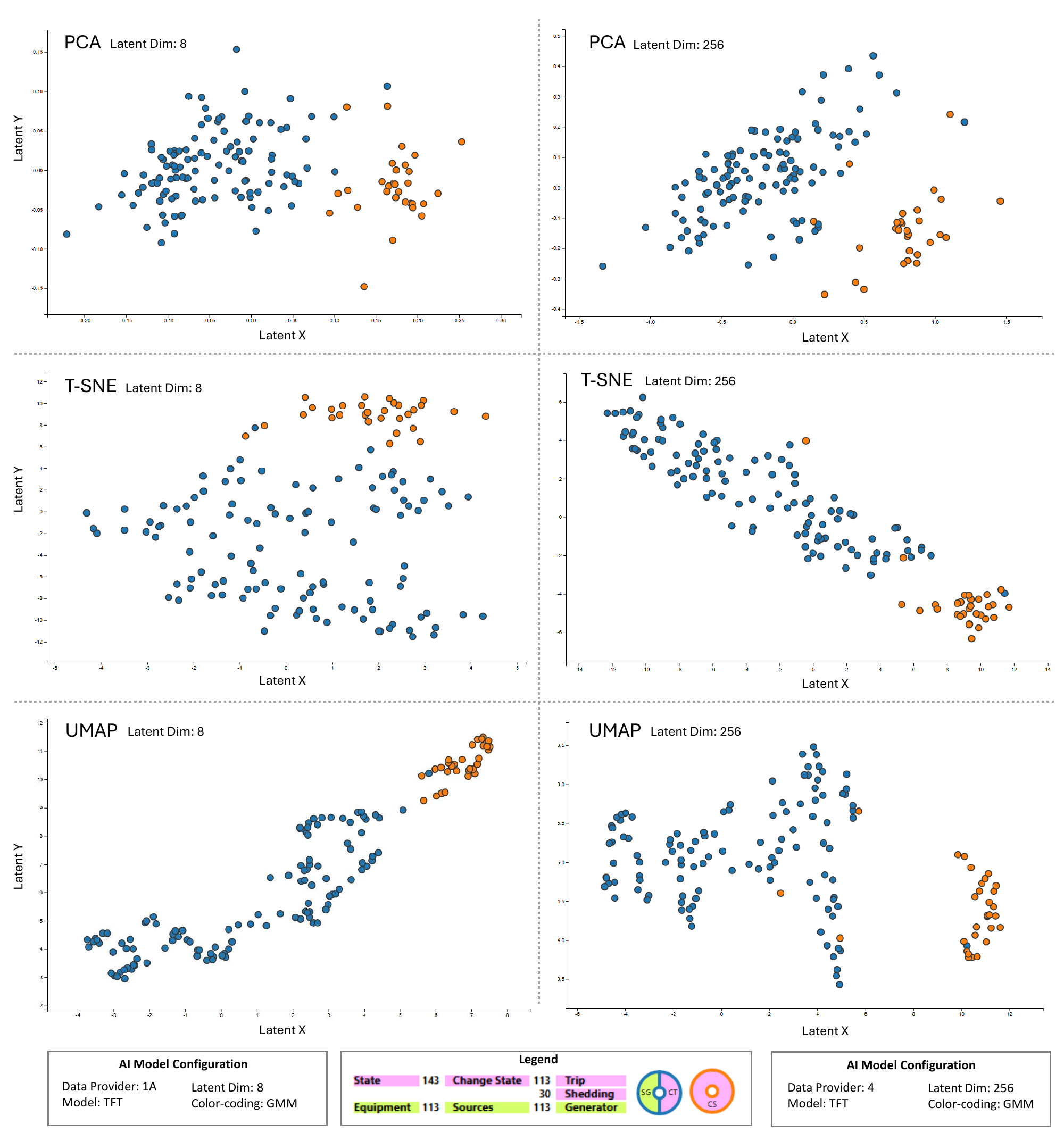}
}
\captionof{figure}{Use case 3: comparing clustering results under different configuration parameters}
\label{fig:usercase-3}

The data-driven exploration of potential interconnections between the signal event patterns EA (Arching), OU (Jumper Failure), and other neighboring event patterns, such as IF (Interrupting Devices - Fuse), DE (Equipment Deterioration), and TN (Nonspecific Transformer Failure), can be also justified using power grid domain knowledge. Arching and jumper failures often serve as precursors or co-occurring events with other equipment-related issues. For instance, Interrupting Devices (IF), such as fuses, are designed to protect the system by isolating faults; their activation may be triggered by arcing or jumper disconnections. Similarly, Equipment Deterioration (DE) and Transformer Failures (TN) are commonly observed consequences of prolonged arcing and mechanical stress from jumper failures, which can accelerate insulation breakdown, thermal damage, and structural degradation. The proximity of these events in the latent space highlights shared underlying causes, such as aging infrastructure, mechanical wear, and transient voltage conditions, reinforcing the importance of identifying these interconnections for improved fault diagnosis and system resilience.

\makebox[\textwidth][c]{
    \includegraphics[width=\textwidth]{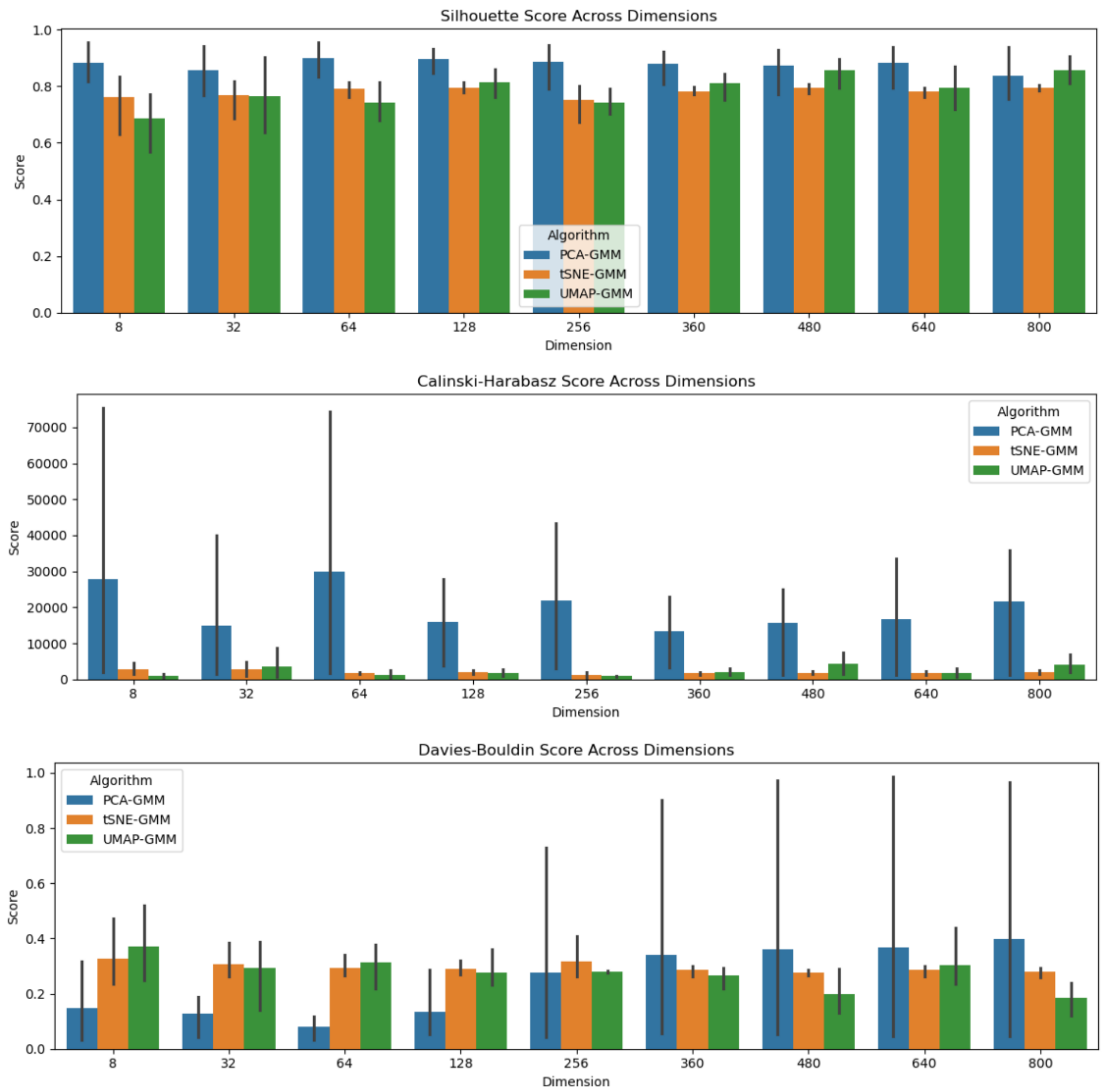}
}
\captionof{figure}{Validating clustering results from TFT, VAE-1D CNN, and VAE-LSTM across different latent dimensions and dimensionality reduction algorithms (PCA, t-SNE, UMAP) using internal validation metrics.}
\label{fig:internal_validation}

\subsubsection{Relative Validation}
\label{subsec:valid-relative}
Based on the methodology described in Section \ref{subsec:Relative Validation}, we employed a correspondence plot to compare the latent representation clusters generated by the TFT and VAE models, as shown in Figure \ref{fig:relative_validation}. The relative validation process evaluates the consistency and alignment of clustering results from these methods by utilizing KNN- and DBSCAN-based cluster similarity metrics. These metrics quantitatively assess the proportion of identical neighbors within clusters, ensuring that only latent vectors with the same cluster labels are considered. For the dataset from Provider 1, applying the t-SNE dimensionality reduction method to latent vectors with a dimensionality of 640 yielded a cluster consistency score of 92.88\% between TFT and VAE-1D CNN models and 93.56\% between TFT and VAE-LSTM models.

In addition to these metrics, the correspondence plot provides a complementary visualization for comparing clustering results. This plot employs Dynamic Time Warping (DTW)-styled lines to connect latent vectors with the same ID across the two methods, offering an intuitive representation of the spatial distributions and structural alignments of clusters. The alignment and length of these lines highlight areas of agreement or divergence between the methods. Consistent clusters are visually distinct and well-aligned, while inconsistencies are localized and easily identifiable, as illustrated in the figure. By combining quantitative similarity metrics with visual analysis, this framework provides a comprehensive and robust approach for comparing and validating the clustering performance of TFT and VAE-based models, offering valuable insights into their reliability and robustness under different configurations. Although the demonstration in Figure \ref{fig:relative_validation} compares the clustering results of TFT and VAE-based models configured with a specific latent dimension and dimensionality reduction method, our methodology is versatile and can be applied to compare latent vector results across different models, latent dimensions, and dimensionality reduction techniques, such as PCA and UMAP. This flexibility enables broader exploratory analyses, allowing users to investigate various configurations and gain deeper insights into model performance and clustering behavior.

\makebox[\textwidth][c]{
    \includegraphics[width=\textwidth]{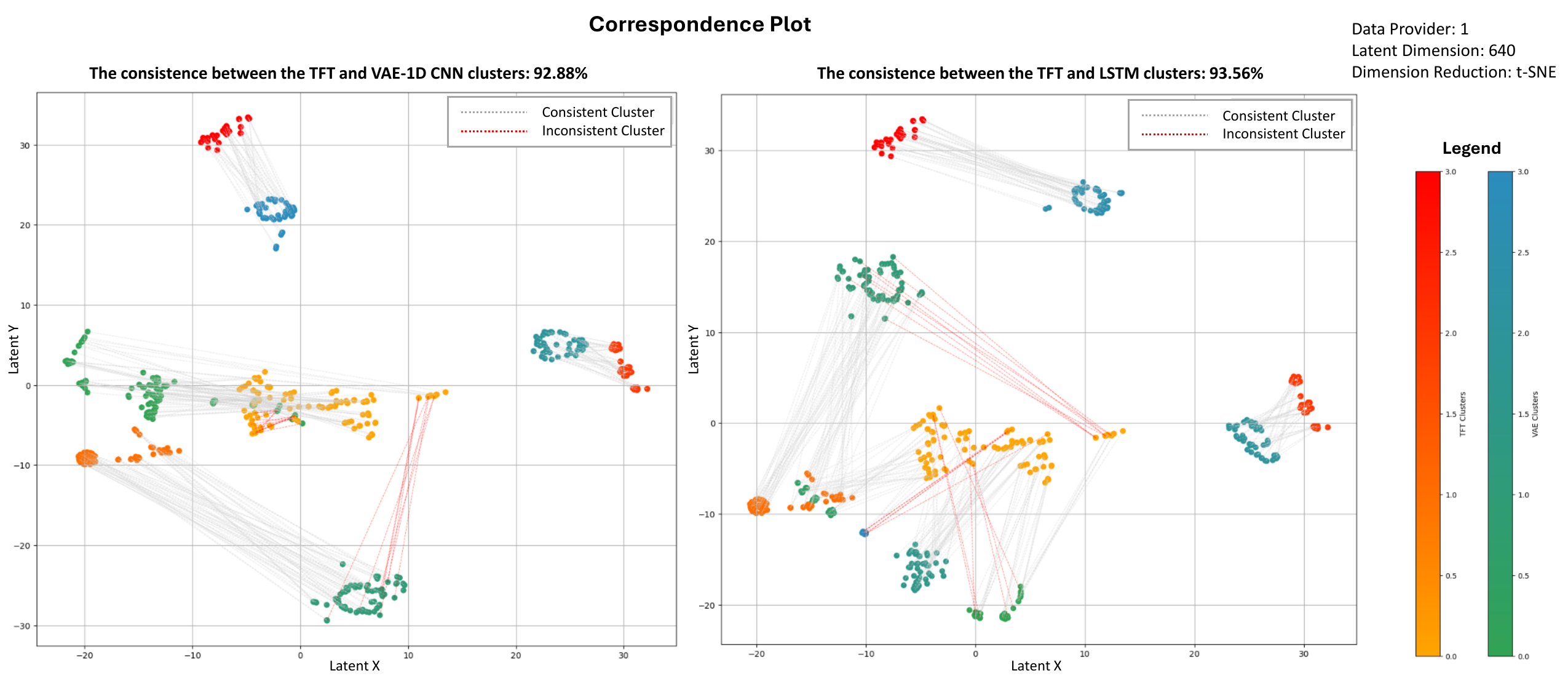}
}
\captionof{figure}{Correspondence plot is developed to facilitate the comparison between the latent representation clusters produced from TFT and VAE models.}
\label{fig:relative_validation}

\subsection{Model Run-time Comparison}

\makebox[\textwidth][c]{
    \includegraphics[width=\textwidth]{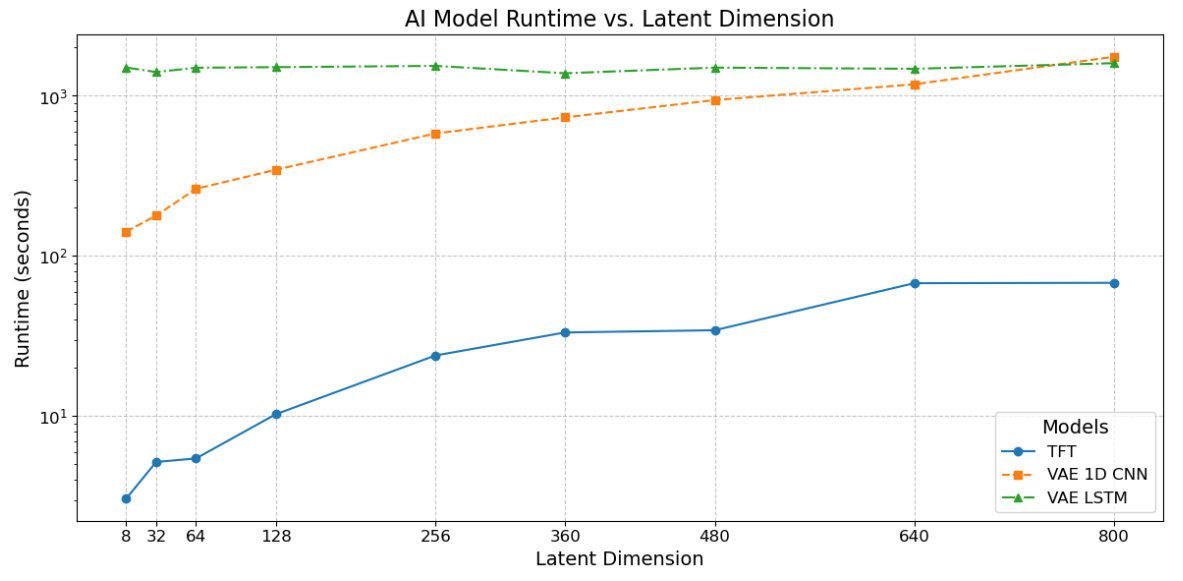}
}
\captionof{figure}{Model run-time comparison between TFT, VAE-1D CNN, and VAE-LSTM across different latent dimensions.}
\label{fig:runtime_vis}

The Model Run-time Comparison evaluates the computational efficiency of TFT, VAE-1D CNN, and VAE-LSTM across varying latent dimensions, as shown in Figure "Model run-time comparison between TFT, VAE-1D CNN, and VAE-LSTM across different latent dimensions". The experiments were conducted on a computing device with the following specifications: 11th Gen Intel(R) Core(TM) i7-1165G7 @ 2.80 GHz (1.69 GHz) processor, 16 GB RAM, and a 64-bit operating system.

The runtime for all models increases with higher latent dimensions, with notable differences in scalability across the three approaches (illustrated in Figure \ref{fig:runtime_vis}). The TFT model demonstrates the fastest runtime, maintaining significantly lower computational costs across all latent dimensions. Even as the latent dimension increases to 800, the runtime remains under 100 seconds, highlighting its computational efficiency. In contrast, the VAE-1D CNN model shows a steady increase in runtime, starting from approximately 100 seconds at lower latent dimensions and reaching over 1000 seconds at higher dimensions. The VAE-LSTM model consistently exhibits the highest runtime, remaining above 1000 seconds across all latent dimensions, reflecting the additional computational overhead of processing sequential dependencies in LSTM-based architectures.

\subsubsection{TFT Efficiency on Time-Series Data}
This comparison underscores the trade-offs between runtime performance and model architecture. While the TFT model offers the most efficient execution, making it ideal for large-scale datasets with short durations (due to its limited attention mechanism) and real-time applications, the VAE-1D CNN and VAE-LSTM models exhibit higher computational costs, particularly as latent dimensions increase. These findings provide valuable insights for users to balance runtime efficiency, model complexity, and task-specific requirements when selecting the appropriate model for their applications. 

\subsubsection{TFT Scalability and Adaptability}
An important observation from our study is that the TFT model demonstrates greater scalability and adaptability for analyzing multivariate time-series data with varying data shapes, such as differing numbers of features and changing time-series durations (data lengths). During our exploratory analysis of the GSEL data, we observed that datasets from different providers (vendors) often exhibit variations in the number of features, sampling rates, and temporal lengths. Unlike VAE-based models, which require significant modifications to the encoder architecture to accommodate data with varying shapes, the TFT model can handle such variations seamlessly without altering its core architecture. This flexibility makes TFT particularly well-suited for heterogeneous datasets, as it can adapt to different data structures while maintaining robust performance. 

\subsection{Limitation and Future Work}
The limitations of our work and their potential solutions include the following:
\begin{description}
    \item [Color-Coding Challenge for Multilabel Data Visualization]: 
    A significant challenge in visualizing multilabel data arises when using color-coding to represent categorical labels in a 2D space. For datasets with a large number of label combinations—potentially exceeding 100 unique combinations—there are not enough distinguishable colors available to adequately represent each combination. As the number of labels grows, it becomes increasingly difficult to assign unique, perceptually differentiable colors to each data point, leading to visual ambiguity, clutter, and reduced interpretability. This limitation complicates the analysis, as users may struggle to identify patterns, relationships, or clusters within the data. Future work could address this challenge by exploring alternative visualization strategies, such as glyph-based representations, where shape, size, or texture encodes additional label information, or interactive filtering techniques that allow users to focus on specific subsets of label combinations. Additionally, techniques like hierarchical grouping or dimensionality reduction of labels can help reduce visual complexity while maintaining meaningful insights.
    \item [Long Time-Sequence Limitation]:
    Our current TFT architecture is not optimized for analyzing time-series data with very long sequences, as the Transformer's attention mechanism can be computationally expensive and may struggle to capture long-range dependencies effectively. We conducted experiments using augmented time-series data from GESL Provider 1, which contains sequences of up to 20K time steps, and observed a noticeable decline in performance based on internal validation metrics. This highlights the limitations of the current TFT implementation for long-sequence modeling due to its quadratic complexity. Future work includes enhancing the TFT by integrating Temporal Convolutional Networks (TCNs) or Recurrent Neural Networks (RNNs), which are well-suited for long-sequence tasks, to improve efficiency and the model's ability to capture long-term temporal dependencies.
    \item [Limitations of External Validation for Multilabel Data]:    
    A key limitation in validating the clustering results of our framework lies in the lack of solid ground truth data for external validation. The dataset is inherently multilabel, where each data point can have multiple class labels, resulting in a large number of unique label combinations. These combinations often imply potential dependencies and interconnections between multiple events, making it difficult to establish a clear, one-to-one mapping between predicted clusters and predefined labels. Standard external validation metrics, such as the F1 score, are designed for single-label classification tasks and do not adequately capture the complexities and interdependencies present in multilabel data. As a result, clustering results cannot be easily validated using traditional quantitative approaches. Instead, visual analytics and domain knowledge play a critical role in interpreting and justifying the clustering outputs, as they allow for the exploration of latent structures, co-occurring labels, and underlying event relationships. Moving forward, the development of domain-specific or hybrid validation techniques that combine quantitative metrics with visual interpretations may provide a more robust solution for assessing clustering performance in multilabel datasets.   

    \item [Visual Encoding Complexity]: A notable limitation of our method lies in the complexity of its visual encodings, which require users to have domain-specific knowledge and invest time in exploring patterns through interactive techniques. While the Multivariate 2D Latent Vector Map and Hierarchical Multi-label Tree provide advanced tools for analyzing multivariate time-series data, their effective use depends heavily on the user’s familiarity with the domain and understanding of the underlying data structures. This requirement may present a barrier to users who lack extensive domain expertise or are unfamiliar with visual analytics workflows. A potential improvement to address this challenge is the incorporation of Large Language Models (LLMs) as AI agents within the framework. These AI agents could serve as visual guides, dynamically explaining complex visual encodings and highlighting key patterns or insights to users. By integrating LLMs to facilitate user understanding and interaction, future iterations of the framework could significantly enhance accessibility, reduce the learning curve, and broaden its applicability to non-expert users.
    \item [Limited Analysis on TFT and VAE Cluster Inconsistency]: The current research introduces a relative validation method using correspondence plots and shape-matching metrics to compare the consistency of latent vector clusters generated by different generative AI models. However, the analysis stops short of a deeper investigation into the causes of cluster inconsistency. Future work could extend this comparative analysis into an ensemble-based approach that characterizes the event patterns contributing to the inconsistencies, focusing on their class labels and data quality. Such an approach could help identify the root causes of disagreements between TFT and VAE-based models, providing valuable insights into underlying data anomalies or model limitations. This deeper understanding would enhance AI explainability, allowing for more transparent and interpretable outcomes in multivariate time-series analysis.
\end{description}

Future work for this study includes the integration of more advanced generative AI technologies to address the current framework's limitations. Specifically, knowledge-augmented large language models (LLMs) could be leveraged to enhance data acquisition, transformation, and pattern justification by utilizing specialized knowledge bases \citep{xu2024leveraging}. This could be enabled through the Retrieval-Augmented Generation (RAG) paradigm, allowing LLMs to perform specialized tasks using domain-specific knowledge \citep{tupayachi2024towards}. Additionally, the generative AI models already employed in the framework, namely the TFT and VAEs, could be further utilized to improve the quality of multivariate time-series data prior to their introduction into the framework for analysis. By refining input data and incorporating domain-driven insights through data augmentation and missing data imputation, these advancements could significantly enhance the framework's robustness and applicability.

\section{Conclusion}
\label{Conclusion}
In this study, we introduced a comprehensive framework that combines advanced generative AI models, namely the Temporal Fusion Transformer (TFT) and Variational Autoencoders (VAEs), with robust visual analytics to facilitate the exploration and interpretation of complex multivariate time-series data. By leveraging dimensionality reduction techniques such as PCA, t-SNE, and UMAP, our framework enables the visualization of latent space representations, offering domain experts intuitive insights into clustering patterns and temporal dependencies. Through case studies utilizing real-world power grid data, we demonstrated the framework's ability to uncover interdependencies among fault events, enhance the imputation of missing class labels, and provide actionable guidance for model parameter tuning.

The results underscore the scalability and efficiency of the TFT model for diverse time-series data shapes, outperforming VAE-based approaches in runtime while maintaining robust analytical capabilities. Moreover, the interactive visual analytics interface significantly enhances human-AI collaboration, enabling domain experts to validate and refine clustering outcomes through human-in-the-loop and relative validation techniques.

Despite its strengths, the framework has limitations, including challenges in visualizing multi-label data with numerous categorical combinations and the absence of comprehensive ground truth datasets for external validation. Future work will address these limitations by developing advanced visual encoding techniques for multi-label data and incorporating external validation mechanisms to further validate clustering reliability.

By bridging the gap between complex data-driven AI techniques and interpretability, this framework advances explainable AI for multivariate time-series analysis, paving the way for more reliable and actionable insights in critical domains such as power grid management and beyond.
\bibliographystyle{elsarticle-harv}
 \bibliography{bib_file}

\end{document}